\documentclass[11pt]{article}
\usepackage{tikz}
\usepackage{amsmath}
\usepackage{amsthm}
\usepackage{amssymb}
\usepackage{amscd}
\usepackage{bbold}
\usepackage{caption}

\usepackage{enumerate}
\usepackage{graphicx}
\usepackage{fancyhdr}
\usepackage{multicol}
\usepackage[margin={1cm,1cm}]{geometry}

\fancypagestyle{firststyle}
{
   \fancyhf{}

}

\begin{document}

\thispagestyle{fancy}
\pagestyle{fancy}

\begin{center}
{\Large{\textbf{Analysis of Sparse Subspace Clustering: \\ Experiments and Random Projection}}}\\
\vspace{.25in}
{\Large{Mehmet F. Demirel }}\\ 
\vspace{.1in}
Department of Computer Science,  University of Wisconsin-Madison  \\ 
\vspace{.2in}
{\Large{Enrico Au-Yeung}} \\ 
\vspace{.1in}
Department of Mathematical Sciences, DePaul University
\vspace{.40in}
\end{center}

\noindent \textbf{\small ABSTRACT \ \ }\ 
Clustering can be defined as the process of assembling objects into a number of groups whose elements are similar to each other in some manner. As a technique that is used in many domains, such as face clustering, plant categorization, image segmentation, document classification,  clustering is considered one of the most important unsupervised learning problems. Scientists have surveyed this problem for years and developed different techniques that can solve it, such as k-means clustering. We analyze one of these techniques: a powerful clustering algorithm called Sparse Subspace Clustering. We demonstrate several experiments using this method and then introduce a new approach that can reduce the computational time required to perform sparse subspace clustering. 
\vspace*{.5in}

\thispagestyle{firststyle}

\begin{multicols}{2}

\section{\textbf{\small Introduction}} 

The notion of clustering data is a significant concept, as it has been used in many fields e.g. medicine, biology, marketing, library systems, etc. It provides us with insight as to what the natural groupings from data look like. For this reason,  it is also considered the most important unsupervised learning problem in the context of machine learning. Furthermore, it can also be used to enhance the accuracy of supervised machine learning algorithms by clustering the data points into similar classes and using these class labels as independent variables in the supervised learning algorithm.

The Sparse Subspace Clustering algorithm can cluster data points that lie in a union of low-dimensional subspaces. The essential idea is that even though we can represent a data point in terms of other points in infinitely many possible ways, selecting a few points from the same subspace yields a sparse representation of that data point.  This technique is capable of dealing with data noise, sparse outlying entries, and missing entries \cite{elhamifar2009sparse} \cite{elhamifar2013sparse}.  It is worth emphasizing one powerful feature of this algorithm is that the number of clusters does not have to be specified or given as an input to the algorithm. 

We implement the  algorithm, then perform several experiments using movie frames from an open-source movie called Elephants' Dream and demonstrate that it is able to group these images based on the similarities among them.  In the context of a movie, the notion of the similarity between these frames refers to whether they are chosen from the similar parts of the movie or not. In general, one can expect that two images that were obtained from within the same 2 seconds of the movie will look alike. 

One movie frame can be represented as one column vector in a matrix $Y$.
The problem is to determine the matrix $C$ so that
$Y = YC$, where the diagonal entries of $C$ are zeros.
Each column of $C$ is a sparse vector.

Lastly, we propose a random projection technique that can dramatically reduce the computational time of the algorithm. We introduce a new random matrix and integrate it into the original optimization problem. We then show that after applying this method, the algorithm's capability of clustering given vectors is not negatively influenced.  As a benefit, the time required to perform the clustering can be reduced.

\section{\textbf{\small Experiments with Sparse Subspace Clustering}} 

We have obtained a number of movie frames from an online open-source movie called Elephants' Dream and used these images in our algorithm to observe its capability to classify them into proper clusters. In these experiments, we chose movie frames from the distinct parts of the movie so that the images in the same group (i.e. the frames from the same part of the movie) would belong to the same cluster (Figure 4).


\makebox[0pt][l]{%
\begin{minipage}{\linewidth}
\centering
\includegraphics[width=.8\linewidth]{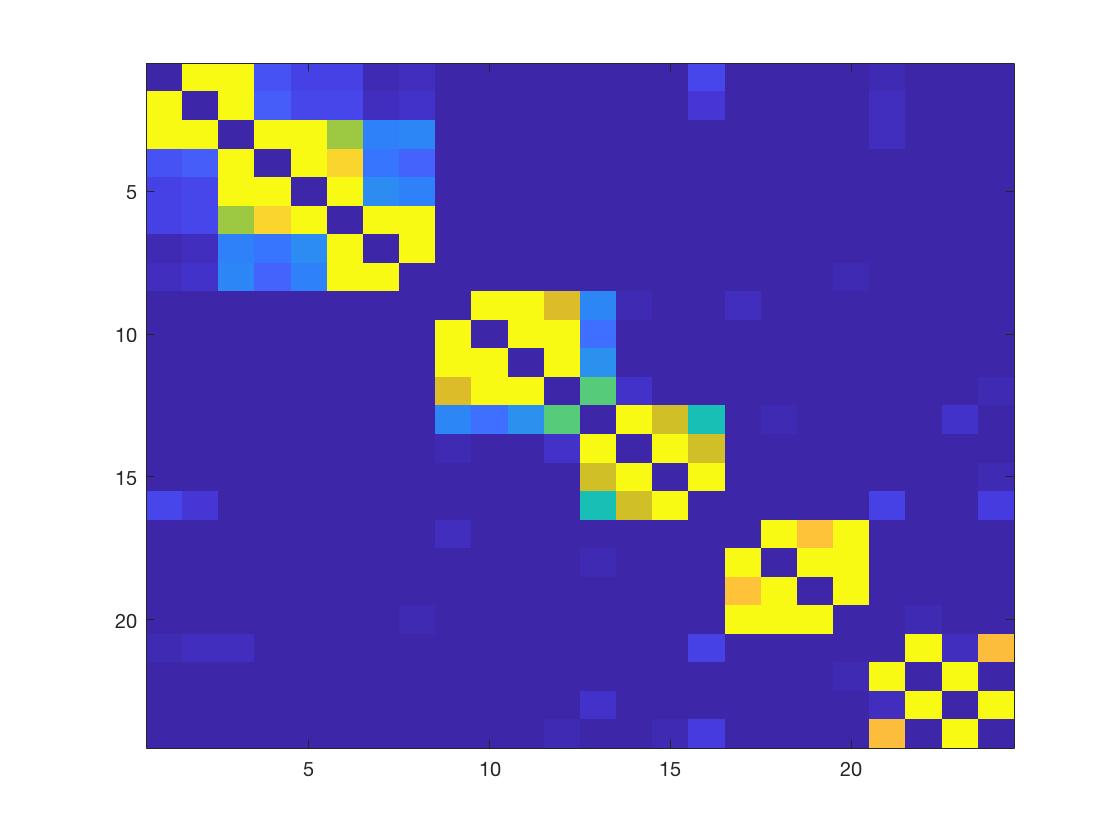}

\captionof{figure}{Visual representation of the resulting W matrix when 24 frames from 3 different parts of the movie used}
\label{fig:graph}
\end{minipage}
}

\makebox[0pt][l]{%
\begin{minipage}{\linewidth}
\centering
\includegraphics[width=.8\linewidth]{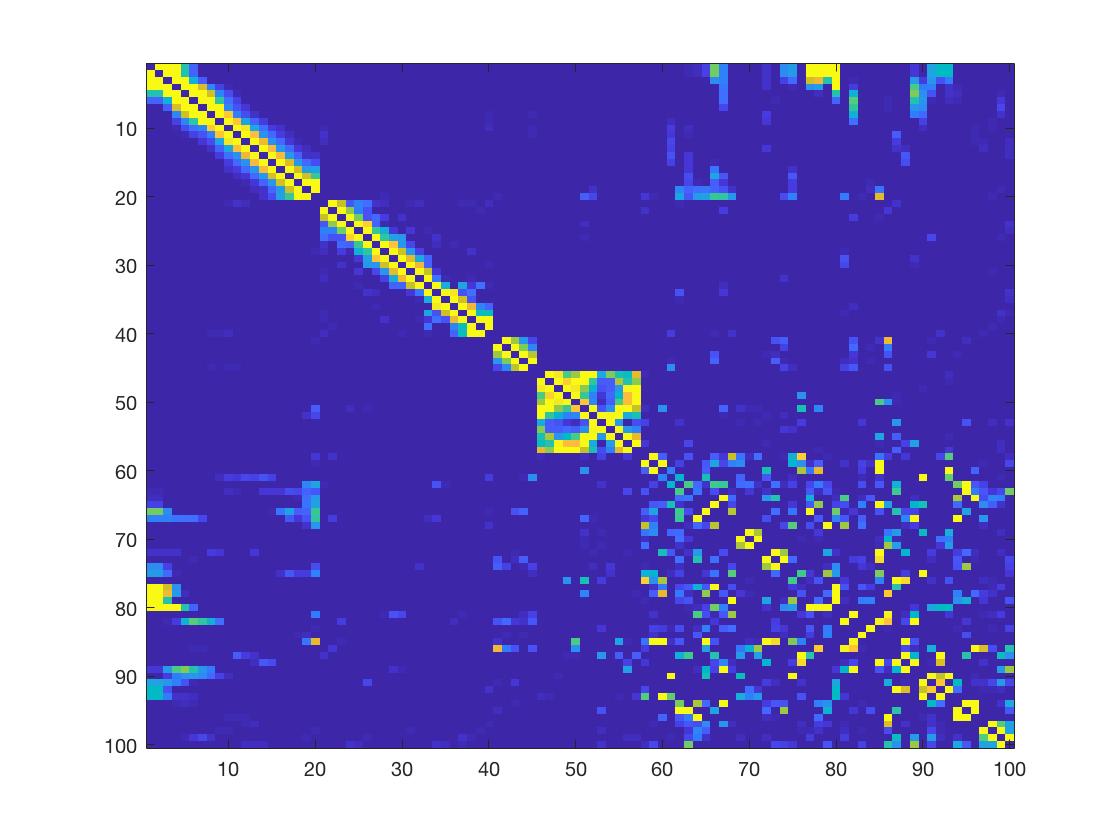}

\captionof{figure}{Visual representation of the resulting W matrix when 100 frames from 5 different parts of the movie used.}
\label{fig:graph}
\end{minipage}
}

 \makebox[0pt][l]{%
\begin{minipage}{\linewidth}
\centering
\includegraphics[width=.8\linewidth]{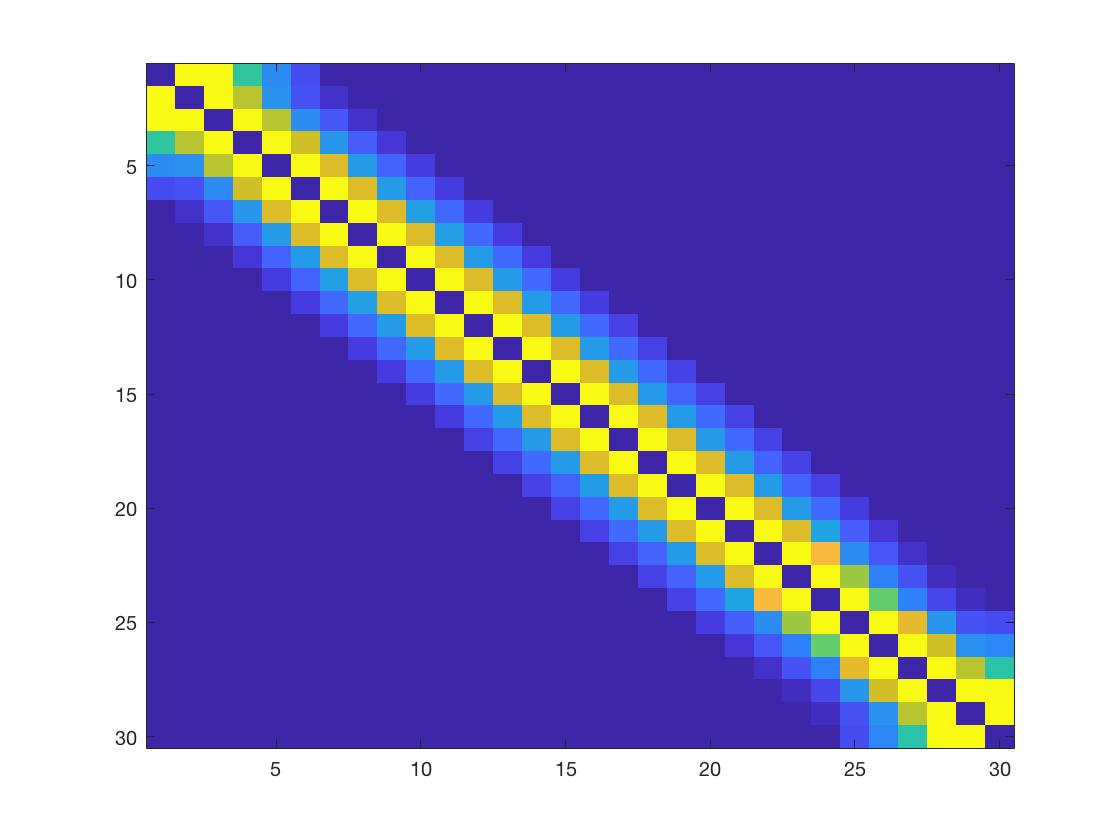}

\captionof{figure}{Visual representation of the resulting W matrix when 30 frames from the same part of the movie used.}
\label{fig:graph}
\end{minipage}
}
\\~\\

\begin{figure*}[!htb]

\centering
\includegraphics[width=.12\textwidth]{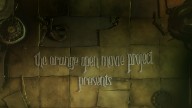}\hfill
\includegraphics[width=.12\textwidth]{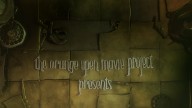}\hfill
\includegraphics[width=.12\textwidth]{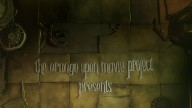}\hfill
\includegraphics[width=.12\textwidth]{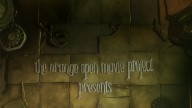}\hfill
\includegraphics[width=.12\textwidth]{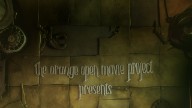}\hfill
\includegraphics[width=.12\textwidth]{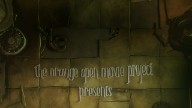}\hfill
\includegraphics[width=.12\textwidth]{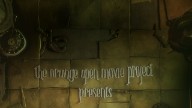}\hfill
\includegraphics[width=.12\textwidth]{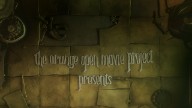}\hfill

\hfill

\includegraphics[width=.12\textwidth]{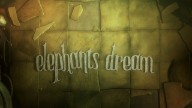}\hfill
\includegraphics[width=.12\textwidth]{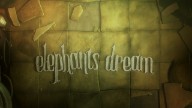}\hfill
\includegraphics[width=.12\textwidth]{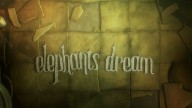}\hfill
\includegraphics[width=.12\textwidth]{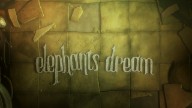}\hfill
\includegraphics[width=.12\textwidth]{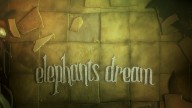}\hfill
\includegraphics[width=.12\textwidth]{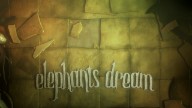}\hfill
\includegraphics[width=.12\textwidth]{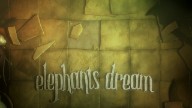}\hfill
\includegraphics[width=.12\textwidth]{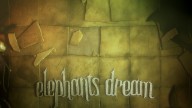}\hfill

\hfill 

\includegraphics[width=.12\textwidth]{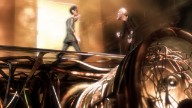}\hfill
\includegraphics[width=.12\textwidth]{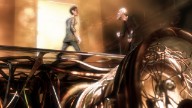}\hfill
\includegraphics[width=.12\textwidth]{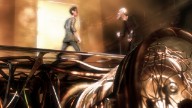}\hfill
\includegraphics[width=.12\textwidth]{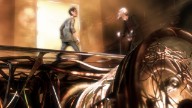}\hfill
\includegraphics[width=.12\textwidth]{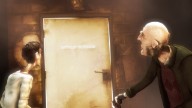}\hfill
\includegraphics[width=.12\textwidth]{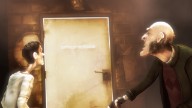}\hfill
\includegraphics[width=.12\textwidth]{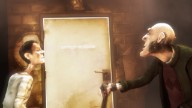}\hfill
\includegraphics[width=.12\textwidth]{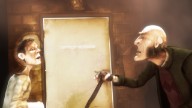}\hfill

\caption{The movie frames used in the experiment whose results are shown in Figure 1. Each row of images forms one cluster since they are  from the same part of the movie.}
\label{images}

\end{figure*}

When taken into consideration, given a number of images, the algorithm was able to identify how each one of these images can be expressed as a linear combination of the others in a sparse way. It is important to note that we did not have to specify the number of clusters in the inputs of the algorithm. While the blueness of pixels in the visual representation of the W matrix represents closeness to zero, the yellowness of the pixels indicates otherwise. For instance, in Figure 1, the algorithm's output demonstrates that the movie frames given to the program could actually be separated into three distinct groups. In fact, the three 8-by-8 yellow sub-squares along the diagonal of the matrix illustrate these three groups.  As an example, the n-th row (or the column) of the resulting matrix demonstrates how the n-th image can be written as a linear combination of the rest of 29 images. 

\section{\textbf{\small Random Projection}} 
The computational time of this algorithm is directly proportional with the size of the matrix Y. Therefore, when the number of the inputs to the algorithm is significantly large and/or the input vectors happen to lie in a very high-dimensional space, the algorithm requires much more time to execute. In order to solve this problem, we decided to use a random Gaussian matrix G so that we can find a representation of Y in a lower dimension. The motivation behind the idea of using this method relies upon the Johnson-Lindenstrauss  lemma \cite{Dasgupta2003}, which states that a small set of points in a high-dimensional space can be mapped into a space of much lower dimension in such a way that the distances between the points are nearly preserved. Therefore, the dimensionality of a data can be reduced in a way that preserves its relevant structure.

The initial optimization problem is
\[ \min \ ||c_i||_1 \ \mbox{ s.t. } \ y_i = Yc_i, \ c_{ii} = 0. \]
Here, $c_i$ is  $i$-th column vector  in matrix $C$. 
       The $i$-th entry of $c_i$ is $c_{ii}$.  
      Minimizing the $l_1$-norm of a column enforces the condition that the vector is sparse (see \cite{omp}, \cite{Tropp2004}, \cite{Tropp2006}).
Let $ \| C \|_1$ be the sum of  the $l_1$-norms of all the colums  in matrix $C$.
The  optimization problem  is equivalent to
\[ \min \ ||C||_1 \ \mbox{ s.t. } \ Y = YC, \ \mbox{diag}(C)=0. \]

\newpage

Accordingly, we have performed several experiments using this technique. As a result, we observed that using a random Gaussian matrix G in the optimization problem does not negatively impact the process of clustering. Rather, it can help reduce the computational time of the algorithm.

\makebox[0pt][l]{%
\begin{minipage}{\linewidth}
\centering
\includegraphics[width=.75\linewidth]{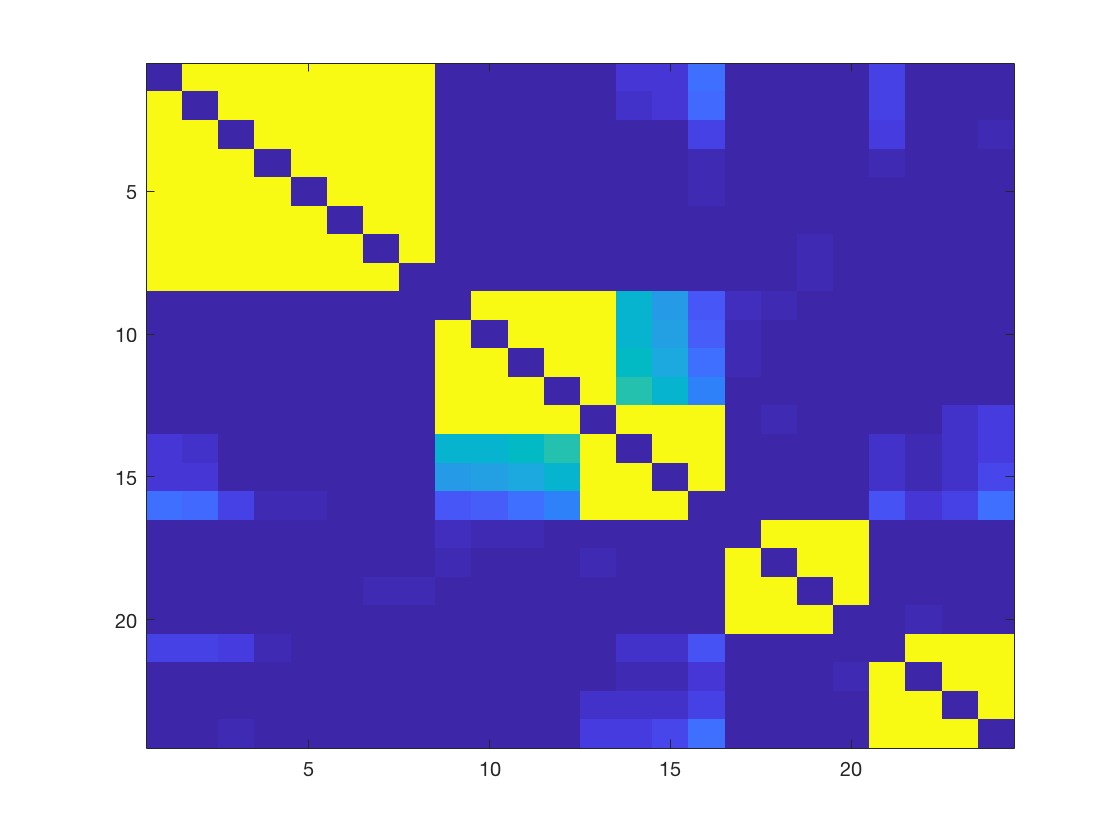}

\captionof{figure}{Visual representation of the resulting W matrix without the use of a random Gaussian matrix, Y is 20736 x 24}
\label{fig:graph}
\end{minipage}
}

\makebox[0pt][l]{%
\begin{minipage}{\linewidth}
\centering
\includegraphics[width=.75\linewidth]{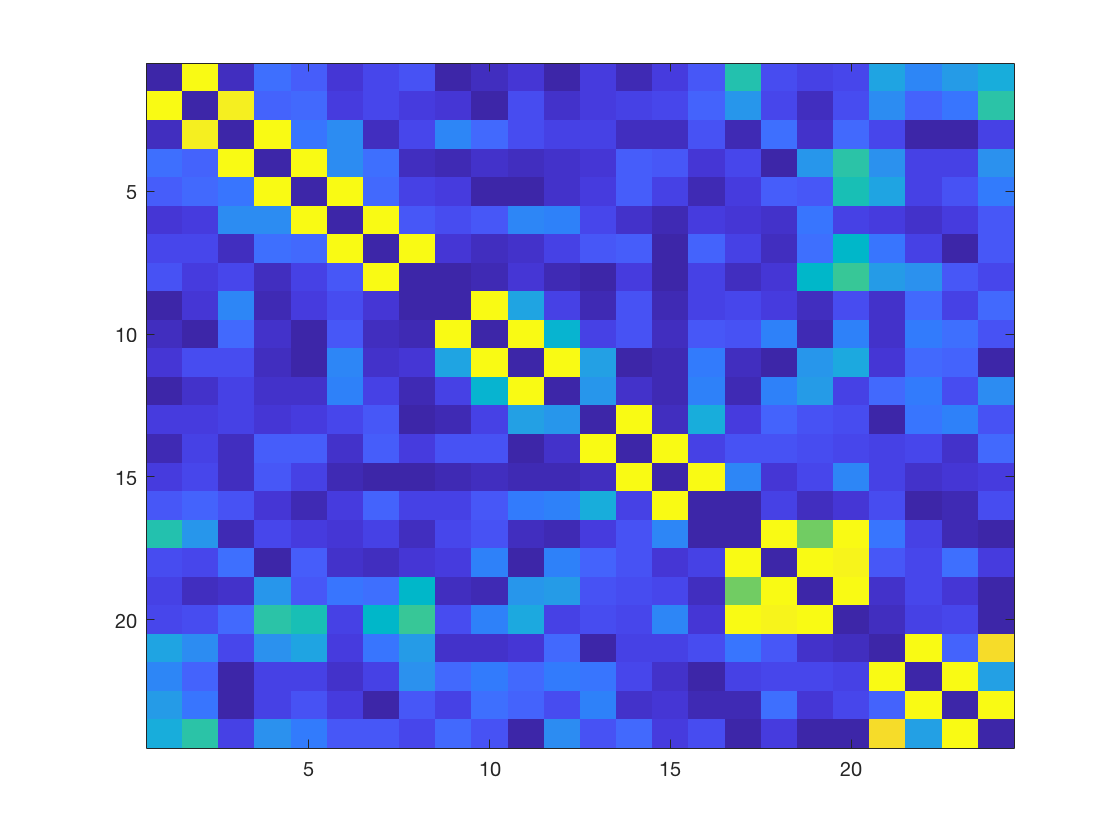}

\captionof{figure}{Visual representation of the resulting W matrix with the Gaussian matrix G (1000x20736)}
\label{fig:graph}
\end{minipage}
}

\makebox[0pt][l]{%
\begin{minipage}{\linewidth}
\centering
\includegraphics[width=.75\linewidth]{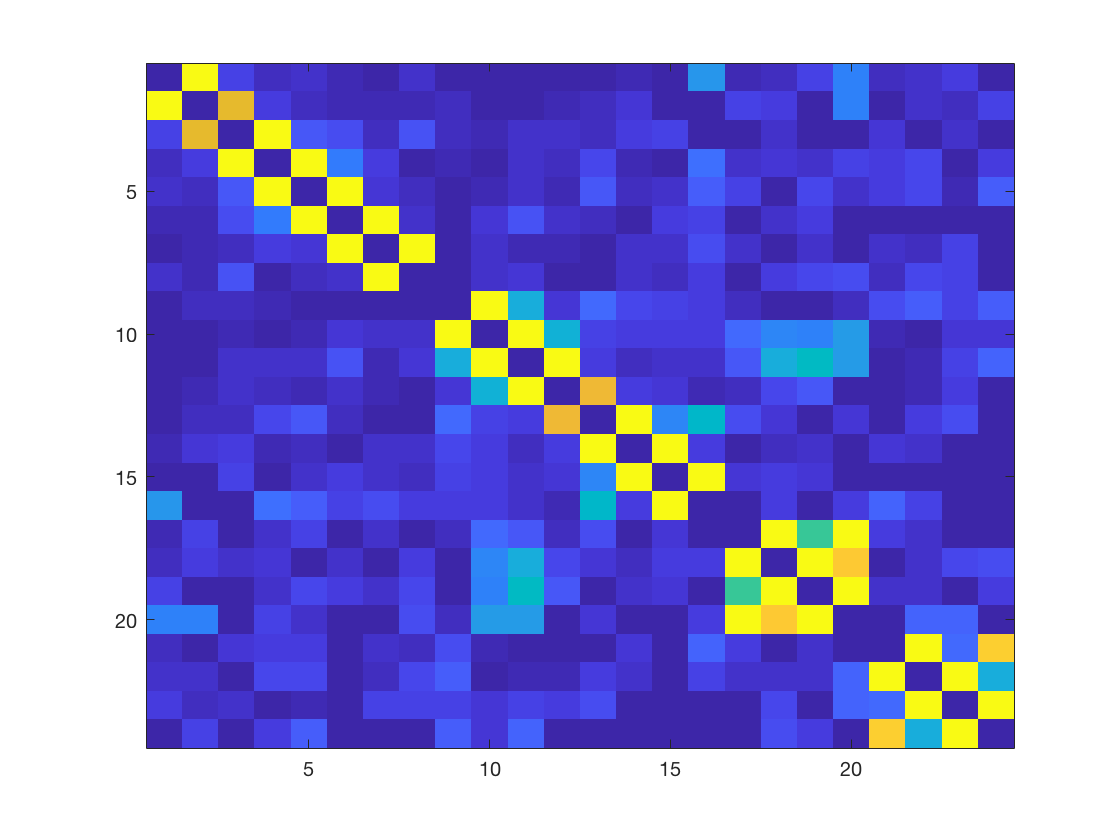}

\captionof{figure}{Visual representation of the resulting W matrix with the Gaussian matrix G (1000x20736)}
\label{fig:graph}
\end{minipage}
}

\hfill

Based on the results of these experiments, we claim that, with the use of a specially-structured random matrix, the computational time of the algorithm can be dramatically reduced. It can be seen that although the resulting W matrix in Figure 1 demonstrates a much clear picture of the three clusters, same clusters can still be observed in the resulting matrices in Figure 5 and 6 as well.

  To implement the algorithm, we consider the  following optimization problem,
\[ \begin{aligned} & \min_{C} \ ||C||_1 +\frac{\mu}{2} \| Y - YC \|_F^2 \\
      & \mbox{ s.t. } \quad C^{T} {\bf{1}} = {\bf{1}}, \quad \mbox{diag}(C) = 0. 
    \end{aligned}\]
    We turn our attention to how to solve this problem  in the next section. 

\end{multicols}

\section{\textbf{\small Optimization Theory}}

    We want to develop an efficient  algorithm to solve the   optimization problem,
\[ \begin{aligned}  & \min_{C} \  ||C||_1 +\frac{\mu}{2} \| Y - YC \|_F^2 \\
      & \mbox{ subject to } \quad  C^{T} {\bf{1}} = {\bf{1}}, \quad \mbox{diag}(C) = 0. 
    \end{aligned}\]
    We first introduce an additional variable, a matrix $A$, and reformulate the problem,
    \[ \begin{aligned}  & \min_{C} \quad  ||C||_1 +\frac{\mu}{2} \| Y - YA \|_F^2 \\
      & \mbox{ subject to } \quad A^{T} {\bf{1}} = {\bf{1}}, \quad A = C - \mbox{diag}(C). 
    \end{aligned}\]
    We next rewrite the objective function by adding two penalty terms.  The objective function is now a convex function.
The problem can be formulated as
 \[ \begin{aligned} 
      \min_{(C, A)} \quad & ||C||_1 +\frac{\mu}{2} \| Y - YA \|_F^2   + \frac{\rho}{2} \| A^{T} {\bf{1}} - {\bf{1}} \|_{F}^2 \\  
      & + \frac{\rho}{2} \| A - \left( C - \mbox{diag}(C) \right)  \|_{F}^2  \\
      & \mbox{ s.t. } \quad A^{T} {\bf{1}} = {\bf{1}}, \quad A = C - \mbox{diag}(C). 
    \end{aligned}\]
    There are two sets of equality constraints, so we introduce a vector $\delta$ and a matrix $\Delta$, which are the Lagrange multipliers. 
     The optimization problem becomes
     \[ \begin{aligned} 
      \min_{(C, A)} \quad & ||C||_1 +\frac{\mu}{2} \| Y - YA \|_F^2   + \frac{\rho}{2} \| A^{T} {\bf{1}} - {\bf{1}} \|_{F}^2 \\  
      & + \frac{\rho}{2} \| A - \left( C - \mbox{diag}(C) \right)  \|_{F}^2  \\
      & + \delta^{T}  \left( A^{T} {\bf{1}} - {\bf{1}} \right) +  \left\langle \ \Delta, A - \left(  C - \mbox{diag}(C) \right) \ \right\rangle. 
    \end{aligned}\]
 The last term in the objective function is an inner product of two matrices.  For any pair of matrices $B_1$ and $B_2$, the inner product is defined by
    \[ \langle B_1, B_2 \rangle = \mbox{Tr}(B_1^{T} B_2). \]
    
    We use an iterative process to solve the optimization problem.  Let $(C_{k}, A_{k})$ be the variables at iteration $k$. Let $(\delta_{k}, \Delta_{k})$ be the Lagrange multipliers at iteration $k$.
    To obtain $C_{k+1}$, we hold the variables $A_{k}, \delta_{k}, \Delta_{k}$ fixed, and minimize the objective function with respect to $C_{k}$.  Then,  $C_{k+1}$ has the closed-form solution,
    \[ \begin{aligned}
    C_{k+1} & = J - \mbox{diag}(J), \\
    J & \equiv \mathcal{T}_{\frac{1}{\rho}}\left( A_{k+1} + \frac{1}{\rho} \Delta_{k} \right),
    \end{aligned} \]
    where the threshold operator $\mathcal{T}_{s}$  is defined by 
    \[ \mathcal{T}_{s}\left(v \right) = \max( |v| - s, 0) \ \mbox{sgn}(v). \]
    While holding $C_{k+1}$ and $A_{k+1}$ fixed, we update the Lagrange multipliers,
    \[ \delta_{k+1} = \delta_{k} + \rho \left( A_{k+1}^{T} {\bf{1}} - {\bf{1}} \right) \]
    and
    \[ \Delta_{k+1} = \Delta_{k} + \rho \left( A_{k+1} - C_{k+1} \right). \]
    
 \section{\textbf{\small Acknowledgements}} 
 
 The first author 
acknowledges the financial support of an Undergraduate Research Assistant Program (URAP) from DePaul University.

\bibliographystyle{plain}
\bibliography{References}





\end{document}